\documentclass[10pt,conference,a4paper]{IEEEtran}

\ifCLASSINFOpdf
\else
\fi

\usepackage[%
  xindy,
  acronym,
]{glossaries}
\glsdisablehyper
\defglsdisplayfirst[acronym]{\emph{#1#4}}

\usepackage[
  backend=bibtex,
   style=ieee,
  citestyle=numeric-comp,
  natbib=true,
  mincitenames=1,
  maxcitenames=1,
]{biblatex}
\bibliography{main.bib}

\usepackage{graphicx}
\usepackage[capitalize]{cleveref} 

\usepackage{subcaption}

\usepackage{tumabbrev}

\newacronym{gl:IoU}{IoU}{intersection over union}
\newacronym{gl:CNN}{CNN}{convolutional neural network}
\newacronym{gl:RNN}{RNN}{recurrent neural network}
\newacronym{gl:FCN}{FCN}{fully convolutional network}
\newacronym{gl:RPN}{RPN}{region proposal network}
\newacronym{gl:CRF}{CRF}{conditional random field}
\newacronym{gl:GAN}{GAN}{generative adversarial network}
\newacronym{gl:DSM}{DSM}{digital surface model}
\newacronym{gl:OSM}{OSM}{OpenStreetMap}
\newacronym{gl:R2UNet}{R2U-Net}{residual recursive U-Net}

\newcommand{\MaskPolydot}{Mask2Poly}
\newcommand{\MaskPoly}{\MaskPolydot\space}

\usepackage{fancyhdr}
\begin{document}

%
\title{Machine-learned Regularization and Polygonization of Building Segmentation Masks}

\author{\IEEEauthorblockN{Stefano Zorzi}
\IEEEauthorblockA{Institute of Computer\\Graphics and Vision\\
Graz University of Technology\\
stefano.zorzi(at)icg.tugraz.at}
\and
\IEEEauthorblockN{Ksenia Bittner}
\IEEEauthorblockA{Remote Sensing \\ Technology Institute, \\German Aerospace Center (DLR)\\
ksenia.bittner(at)dlr.de}
\and
\IEEEauthorblockN{Friedrich Fraundorfer}
\IEEEauthorblockA{Institute of Computer\\Graphics and Vision\\
Graz University of Technology\\
fraundorfer(at)icg.tugraz.at}}

\maketitle

\lhead{\centering This manuscript was accepted to be presented at the IEEE International Conference of Pattern Recognition 2020.}
\thispagestyle{fancy}

\begin{abstract}
We propose a machine learning based approach for automatic regularization and polygonization of building segmentation masks.
Taking an image as input, we first predict building segmentation maps exploiting generic \textsc{\gls{gl:FCN}}. 
A \textsc{\gls{gl:GAN}} is then involved to perform a regularization of building boundaries to make them more realistic, \ie having more rectilinear outlines which construct right angles if required. 
This is achieved through the interplay between the discriminator which gives a probability of input image being true and generator that learns from discriminator's response to create more realistic images.
Finally, we train the backbone \textsc{\gls{gl:CNN}} which is adapted to predict sparse outcomes corresponding to building corners out of regularized building segmentation results. 
Experiments on three building segmentation datasets demonstrate that the proposed method is not only capable of obtaining accurate results, but also of producing visually pleasing building outlines parameterized as polygons.
\end{abstract}

\IEEEpeerreviewmaketitle
\glsresetall
\section{Introduction}
The ability to extract vector representations of building polygons from aerial or satellite imagery has become a hot topic in numerous remote sensing applications, such as urban planning and development, city modelling, cartography, \etc. 
The interest in and the development of new methodologies was also motivated by the current existence of several public benchmark datasets, like INRIA~\cite{maggiori2017dataset}, SpaceNet~\cite{Urban3D2017}, and CrowdAI~\cite{Mohanty:2018}. 
The classical approaches in this research field mostly focused on the assignment of the semantic class to each pixel in the image, obtaining classification masks as output~\cite{tokarczyk2013beyond,yuan2016automatic,bittner2018building,paper1}.
However, for many applications, the more advanced output in form of vector information is under demand.
In this work, we aim to provide not only building segmentation results, which outlines follow the realistic building forms, mainly straight lines and right angles, but also to generate a  polygonal vector structure for each building instance. 

\Glspl{gl:CNN} have brought significant contributions to the field of computer vision, establishing themselves as the basis of semantic and instance segmentation.
However, while performing the pixel-wise classification with high accuracy, they have problems with delineating the exact and regular building boundaries.
To overcome this issue, we apply geometry constraints in the pixel domain using an adversarial loss to regularize the boundaries. 
Specifically, the generative part of the proposed \gls{gl:GAN}-based architecture takes as input the segmentation results obtained from \gls{gl:R2UNet} or the ideal segments from the dataset's ground truth. 
By getting the gradient feedback from the discriminator which task is to verify if its input comes either from regularized segmentation mask or ideal one, the generator learns to output the improved outline contours of our initial segmentation. 

\begin{figure}
\centering
\includegraphics[width=1\linewidth]{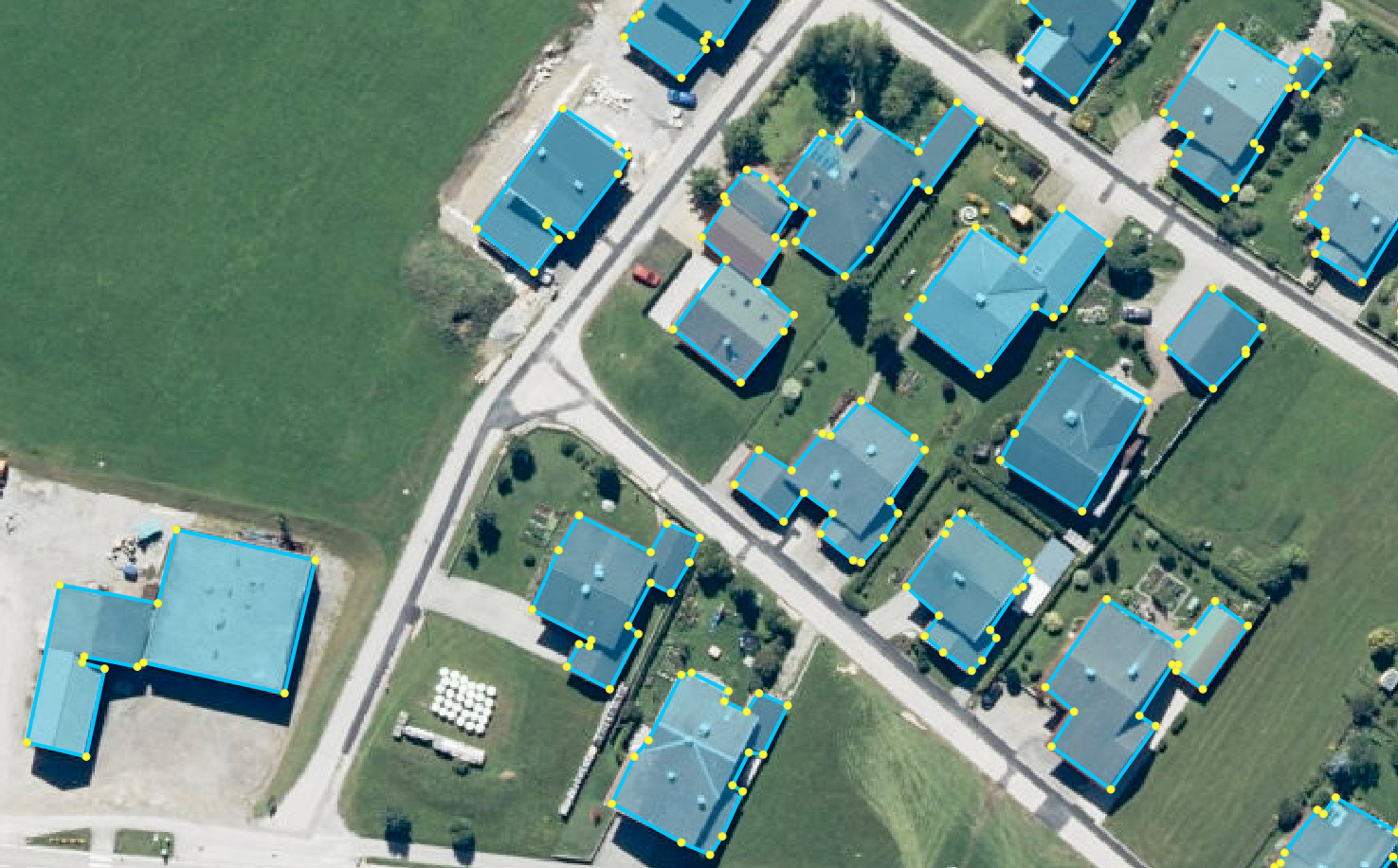}
\caption{Building polygon results from our proposed methodology overlaid on top of a sample area from the Inria dataset.}
\label{fig:firstpage}
\end{figure}


In the literature, several methodologies have already made an attempt to directly predict vertices of object boundaries using \gls{gl:CNN} paradigm.
They are either based on iterative prediction of outline points for one object at a time~\cite{castrejon2017annotating,acuna2018efficient} with possible interaction by users for corrections, or predicting only 4-sided polygons~\cite{girard2018end}.
However, real world buildings are not constrained to a certain amount of corners.  
Motivated by this ideas,~\citet{li2019topological} proposed a \gls{gl:RNN} above the \gls{gl:RPN} which step by step predicts the possible corners for a single building within every region of interest.
In our method, we do not want to be limited to corners prediction for a single building centered inside the input patch. 
The proposed \MaskPoly network is trained to predict an arbitrary number of corners (depending on structure complexity) for random number of buildings in the image scene from the regularized segmentation results. 
Some results of polygonal representations after obtaining the corner predictions from \MaskPoly are shown in \cref{fig:firstpage}.

In~\cref{sec:relatedwork}, we review state-of-the-art methodologies in the related field. 
The details of designed architectures and the intuition behind selected objective functions are then presented in~\cref{sec:method}.
In~\cref{sec:Experiments}, we demonstrate the effectiveness showing qualitative and quantitative results of our approach on three publicly available datasets, \ie INRIA~\cite{maggiori2017dataset}, SpaceNet~\cite{Urban3D2017} and CrowdAI~\cite{Mohanty:2018}.
\cref{sec:Conclusion} concludes the paper.

\begin{figure*}[thbp]
\centering
\includegraphics[width=1.0\linewidth]{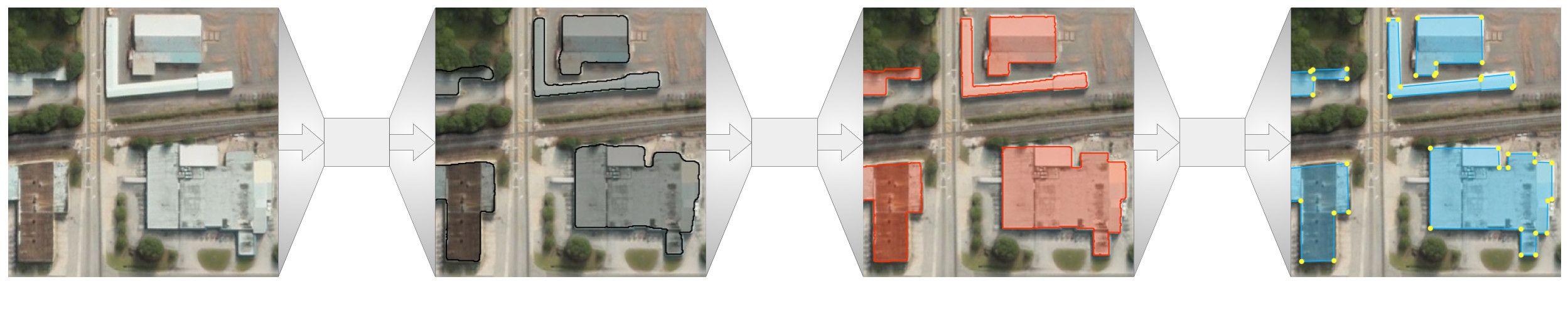}
\put (-75,6) {\scriptsize{Extracted polygons}}
\put (-210,6) {\scriptsize{Regularized mask}}
\put (-355,6) {\scriptsize{Segmentation mask}}
\put (-490,6) {\scriptsize{Input image}}
\put (-405,58) {\scriptsize{SEG}}
\put (-265,58) {\scriptsize{REG}}
\put (-124,58) {\scriptsize{M2P}}
\caption{The schematic overview of the proposed pipeline for automatic extraction of regularized building polygons. Buildings are initially detected and segmented by a \gls{gl:FCN} (result shown in black). A footprint regularization network is then applied to the segmentation mask in the pixel domain (red). Finally, building polygons are extracted from the regularized mask (cyan, vertices highlighted in yellow).}
\label{fig:pipeline}
\end{figure*}
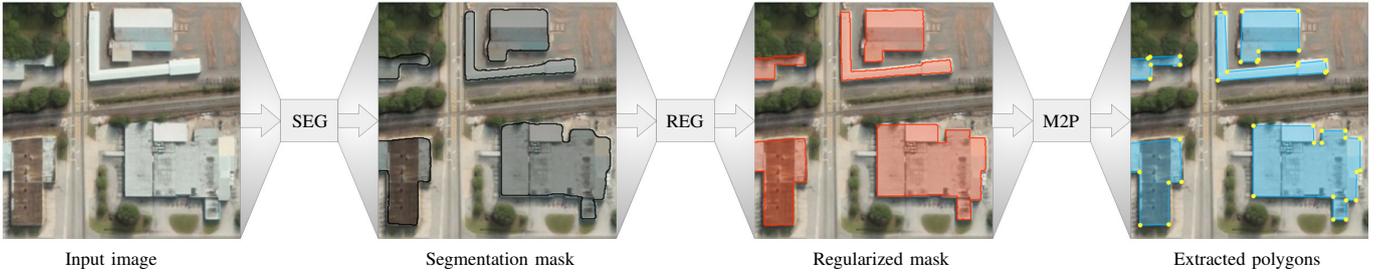

\section{Related work}
\label{sec:relatedwork}

\textbf{Building segmentation} from top view images has been one of the main research topics in remote sensing for decades. 
Before the deep learning era, the traditional methodologies for building footprint extraction relied on multi-step workflows utilizing detected low-level features to form building hypotheses~\cite{huertas1988detecting,guercke2011building}, assumptions that buildings compose of regular rectangular shapes~\cite{kim1999uncertain,bredif2013extracting} and similarities of spectral reflectance values between building appearances\cite{huang2012morphological,baluyan2013novel}.
After the introduction of more powerful hardware, recent approaches began to heavily utilize deep convolutional networks for automatic building delineation providing state-of-the-art  results.
The task is approached via pixel-wise semantic segmentation applying \glspl{gl:FCN} on satellite or airborne images using the benefit of their high-resolution spectral information~\cite{yuan2016automatic,hamaguchi2018building}.
Some methodologies embedded additional information in forms of heights from \glspl{gl:DSM}~\cite{lagrange2015benchmarking,bittner2018building} or \gls{gl:OSM}~\cite{audebert2017joint} together with the spectral information to increase the evidence of buildings.

In the last few years, UNet-based architectures became one of the most successful models for segmentation and detection tasks not only in medical images but also in remote sensing.
Motivated by recently proposed UNet-based models that achieved state-of-the-art performances in different building extraction challenges~\cite{iglovikov2018ternausnetv2, hamaguchi2018building} , the variant of UNet with residual and recurrent layers~\cite{alom2018recurrent} is utilized in this work.

\textbf{Building segmentation regularization} has been getting increased attention over the recent years.
Because neural networks try to decide for each image pixel whether it belongs to a building or not, they do not consider its geometry. 
As a result, building segmentation results have very often a blob-like appearance. 
Therefore, a footprint regularization step is very important to enforce that the resulting outlines not only match the ground truth but also have realistic appearances. 
\Citet{zhao2018building} proposed to regularize building instances obtained from semantic segmentation networks applying multi-step polygon simplification methods. 
\Citet{marcos2018learning} proposed a more advanced architecture by integrating the classic active contour model of~\citet{kass1988snakes} into deep \gls{gl:CNN} to perform a joint end-to-end learning. 
In the following work, \citet{cheng2019darnet} introduced a network based on a polar representation of active contours which prevent self-intersections and enforces outlines to be even closer to the ground truth.   
Work most related to ours is \citet{paper1}, which looked at the problem differently.
The authors of this paper trained the regularization network in an unsupervised manner using adversarial losses together with Potts~\cite{tang2018regularized, tang2018normalized} and normalized cut~\cite{tang2018normalized} regularization losses which embedded additional knowledge about building boundaries from the intensity image to the network. 
In our work, we extend the algorithm proposed in \cite{paper1} redefining the training procedure and the architecture of the regularization network to obtain better results both in qualitative and quantitative terms.

\textbf{Polygon prediction} is a difficult but crucial step for multiple disciplines as it provides vector-based data representations.
Typically, semantic segmentation results are vectorized employing Douglas-Peucker~\cite{douglas1973algorithms}, RANSAC~\cite{fischler1981random} or Hough transform~\cite{duda1971use} algorithms as a post-processing step. 
Recent approaches made an attempt to integrate a vectorization procedure into an end-to-end deep learning-based model. 
The approach of~\citet{castrejon2017annotating} and the followed work of ~\citet{acuna2018efficient} sequentially produce polygonal vertices around the object boundary based on \gls{gl:RNN}. Although these methodologies provided impressive results, they are different from our proposed algorithm in terms of the size and amount of polygonized objects (an image crop containing only one object is annotated per procedure).
Moreover, a human annotator's interaction is allowed during the prediction of polygonal vertices to correct them if needed. 
In contrast, we propose a deep learning-based methodology which automatically predicts polygon vertices without any limitation on the amount of objects within an input image.

\section{Proposed method}
\label{sec:method}

In this paper, we propose a pipeline for building extraction that not only aims to achieve state-of-the-art segmentation accuracy, but also tries to predict visually pleasing building polygons.

The pipeline is composed by three consecutive and independent steps.

As a first step, a \gls{gl:FCN} is used to detect and segment building footprints given an intensity image.
The resulting segmentation can achieve great accuracy in terms of \gls{gl:IoU}, recall and completeness, but the predicted building boundaries do not have a regular shape since there are no constraints on the building geometry.

In order to produce a more realistic segmentation, we further refine the result through a second \gls{gl:CNN} trained using a combination of adversarial, reconstruction and regularized losses.
As a result, the extracted building footprints have a more regular shape, with sharp corners and straight edges.
As we show later in \cref{sec:Experiments}, this step greatly increases the footprints quality without losing segmentation accuracy.

Finally, we extract a polygon for each building instance detecting the corners from its regularized mask.

In the subsequent sections, we describe in more detail each component of the pipeline.

\subsection{Building detection and segmentation}
The first step in the proposed method aims to detect and outline the boundaries of the buildings present in the satellite or aerial image.
This task can be solved exploiting one of the many instance or semantic segmentation networks proposed in literature, trained using cross-entropy losses.
Since the three stages of the pipeline are independent from each other, it is possible to choose the instance of semantic segmentation network which is best suited or which performs best on the specific dataset.
In this work, we decided to use as segmentation baseline the \gls{gl:R2UNet} proposed in \cite{alom2018recurrent}, a simple but yet precise network which guarantees high building segmentation accuracy.

\subsection{Regularization of the segmentation}
\begin{figure*}[thbp]
\centering
\includegraphics[width=0.95\linewidth]{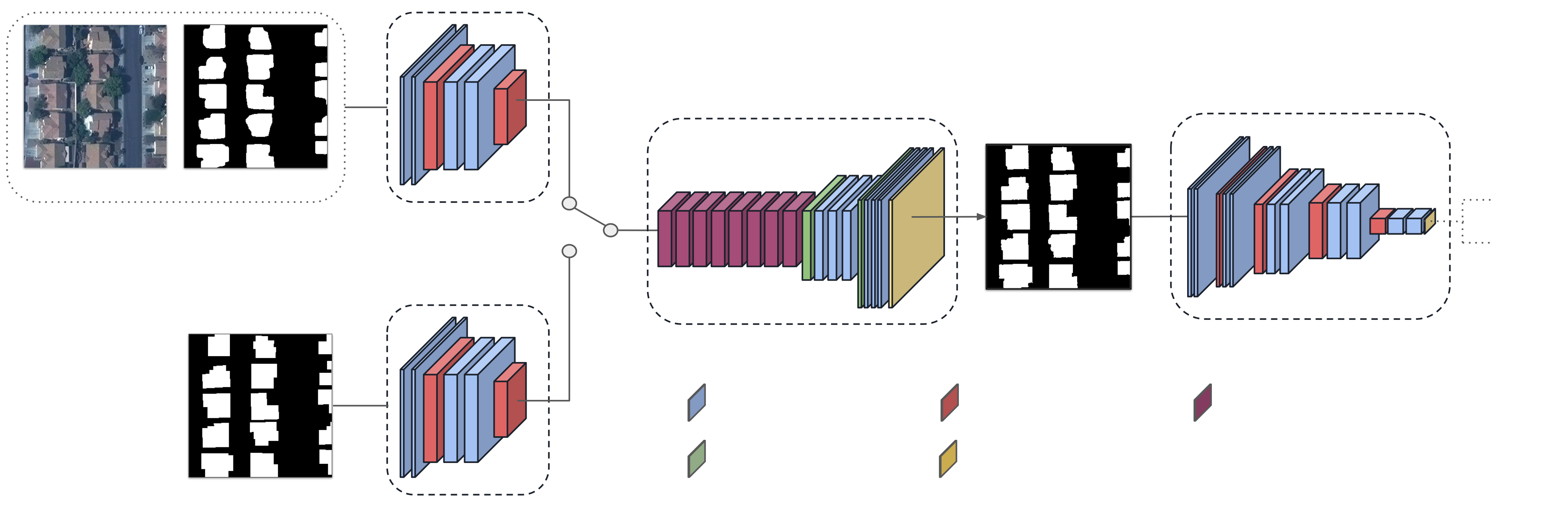}
\put (-188,15) {\scriptsize{conv 1$\times$1}, sigmoid}
\put (-188,32) {\scriptsize{max pool 2$\times$2}}
\put (-266,28) {\scriptsize{conv 3$\times$3},}
\put (-266,34) {\scriptsize{batch norm, ReLU}}
\put (-266,15) {\scriptsize{up-sampling 2$\times$2}}
\put (-188,61) {\scriptsize{Either regularized or}}
\put (-186,53) {\scriptsize{reconstructed mask}}
\put (-108,32) {\scriptsize{residual layer}}
\put (-468,99) {\scriptsize{Image}}
\put (-429,99) {\scriptsize{Segmentation}}
\put (-424,2) {\scriptsize{Ideal mask}}
\put (-465,156) {\textbf{$z$}}
\put (-410,156) {\textbf{$x$}}
\put (-410,58) {\textbf{$y$}}
\put (-335,100) {\textbf{$E_G$}}
\put (-335,9) {\textbf{$E_R$}}
\put (-282,110) {\textbf{$F$}}
\put (-50,110) {\textbf{$D$}}
\put (-22,94) {\scriptsize{$true$}}
\put (-22,81) {\scriptsize{$false$}}
\caption{Workflow of the proposed regularization framework. It is composed of two paths: the generator path ($E_G \rightarrow F$) produces the regularized building footprint mask; the reconstruction path ($E_R \rightarrow F$) encodes and decodes the ideal input mask ensuring to have the same real valued masks as input to the discriminator.}
\label{fig:workflow}
\end{figure*}
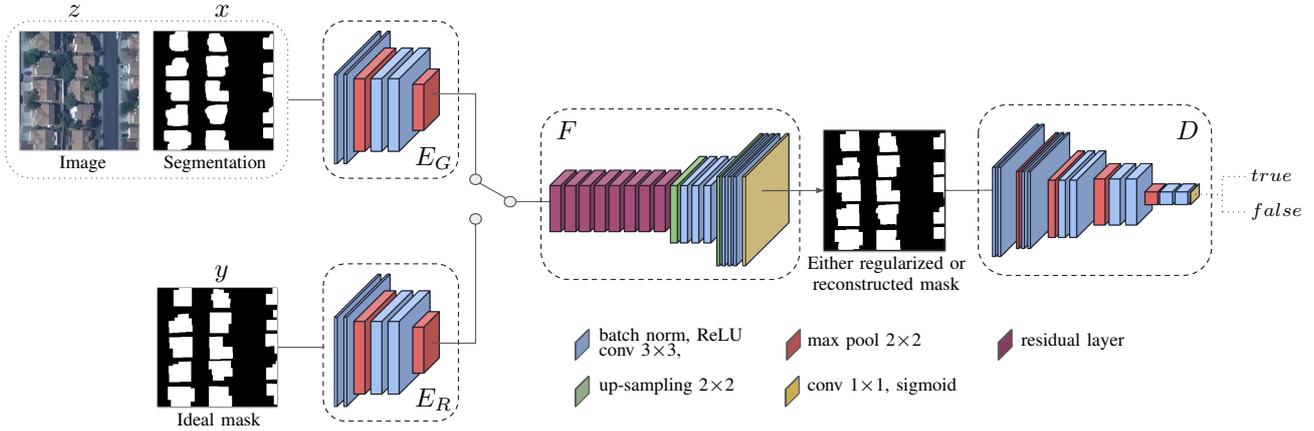

The footprints predicted by the segmentation network typically have rounded corners and irregular edges due to the lack of geometric constraints during the prediction.
Extracting building polygons from the initial building segmentation is a hard task that could lead to errors in the corners proposal procedure.
For this reason, as a second step, we use a \gls{gl:CNN} for building regularization that aims to produce building footprints with regular and visually pleasing boundaries.

This translation can be successfully achieved training a \gls{gl:GAN} network composed by two different models.
One of these networks is a generator which tries to generate a regularized version of the segmentation mask and the other network is a discriminator that examines generated and ideal footprints and estimates whether they are real or fake.
The goal of the generator is to fool the discriminator, and as both networks get better and better at their job over the training, eventually the generator is forced to generate building footprints which become more realistic with each iteration.

The generator aims to learn a mapping function between the domain $X$, composed by segmented footprints, and the domain $Y$, made of ideal footprints, given the training samples $\{x_i\}^N_{i=1}$ where $x_i \in X$ and $\{y_i\}^M_{i=1}$ where $y_i \in Y$.
To further improve the results we also exploit the intensity images, $\{z_i\}^N_{i=1}$ where $z_i \in Z$, training the model with an additional regularized loss.

The generator performs the regularization $G : \{ X, Z \} \xrightarrow{} Y$ exploiting a residual autoencoder structure, as shown in \cref{fig:workflow}.

The regularized footprint is produced through the path composed by the encoder $E_G$ and the residual decoder $F$, so the generator $G$ can be seen as their combination $G(x,z) = F(E_G(x,y))$.

The discriminator network $D$ tries to estimate whether the presented images are regularized footprints, generated by $G$, or ideal ones.
The reason behind this path is to derive a reconstructed version of $y$.
However, the adversarial network can easily distinguish two distributions, since the ideal mask is one-hot encoded with zeros and ones and the output of the autoencoder can range between zero and one. 
Therefore, both reconstructed and regularized image samples are generated using the same network $F$.
Due to the joint training of two autoencoders with the common decoder, the proposed architecture is ensured to be stable and, as a result, escapes the situation where the discriminator wins.

\subsubsection{Objective Function}


Three types of loss functions in the learning procedure are used motivated by the good building footprints produced in \cite{paper1}: \textit{adversarial loss}, \textit{reconstruction losses} and \textit{regularized loss}.

The \textit{adversarial loss}, introduced in~\cite{goodfellow2014generative}, is used to learn the mapping function between the domain $X$ and $Y$, encouraging the generator $G$ to produce footprints similar to the ideal samples.
This component of the objective function acts as a constraint for the  geometry boundaries of the buildings and it is expressed as:
\begin{equation}
\label{eqation:loss_gan_G}
\begin{split}
\mathcal{L}_{GAN}(G,D) = {E}_{x,z}[\log(1-D(G(x,z))]
\end{split}
\end{equation}

The discriminator $D$ is trained to distinguish regularized and reconstructed footprints and its objective function can be expressed as:
\begin{equation}
\label{eqation:loss_gan_D}
\begin{split}
\mathcal{L}_D(G,R,D) &= {E}_{y}[\log(1-D(R(y)))] 
\\&+ {E}_{x,z}[\log D(G(x,z)]
\end{split}
\end{equation}

where the path $R(y)=F(E_R(y))$ encodes and reconstructs the ideal mask and the path $G(x,z)=F(E_\text{G}(x,z))$ generates the regularized footprints.

The \textit{reconstruction} term is introduced to force the generator $G$ to produce building footprints having an overall shape and pose similar to the segmentations received as input.
The loss is also computed through the reconstruction path $R$ to obtain a reconstructed version of the ideal mask.
As reconstruction loss we simply use \textit{binary cross entropy} and two losses can be written as:
\begin{equation}
\begin{split}
\mathcal{L}_{rec_G}(G) &= -{E}_{x,z} [x \cdot \log G(x,z)]
\\
\mathcal{L}_{rec_R}(R) &= -{E}_{y} [y \cdot \log R(y)]
\end{split}
\end{equation}

Alongside the adversarial and regularized losses, a soft version of the Potts and Normalized Cut criterions are used to exploit the information of the intensity image to further improve the regularization results.
The Potts and the Normalized Cut methods are popular graph clustering algorithms originally proposed for image segmentation. 
As demonstrated in \cite{paper1}, these terms can be effectively minimized by the generator $G$.
As a result, the final footprints are aligned to the building boundaries observed in the intensity image.

The Potts and the normalized cut losses can be expressed as:
\begin{equation}
\label{eq:regularized}
\begin{split}
\mathcal{L}_{Potts}(G) = {E}_{x,z} \sum_{k}^{} S^{k\top} W (1-S^k)
\\
\mathcal{L}_{ncut}(G) = {E}_{x,z} \sum_{k}^{} \frac{S^{k\top} \hat{W} (1-S^k)}{1^\top \hat{W} S^k}
\end{split}
\end{equation}

where $S = G(x,z)$ is the k-way softmax mask generated by the network and $S^k$ describes the vectorization of its $k$-th channel.
$W$ and $\hat{W}$ are matrices of pairwise discontinuity costs and each term describes the weight between two nodes (or pixels) and it is computed using a gaussian kernel over the RGBXY space.

The full objective used to jointly train the generator path $G$ and the reconstruction path $R$ is a linear combination between the adversarial loss, the regularized loss and the reconstruction losses.
\begin{equation}
\label{eq:full_objective}
\begin{split}
\mathcal{L}_{}(G,R,D) &= \alpha \mathcal{L}_{GAN}(G,R,D)\\
 &+ \beta \mathcal{L}_{rec_G}(G) + \gamma \mathcal{L}_{rec_R}(R) \\
 &+ \delta \mathcal{L}_{Potts}(G) + \epsilon \mathcal{L}_{ncut}(G)
\end{split}
\end{equation}

It's worth noting that these loss components are obtained by connecting the encoders $E_R$ and $E_G$ to the residual decoder $F$ one at a time.
Once the full objective is computed, $E_G$, $E_R$ and $F$ are updated jointly.

\subsection{Polygon extraction}
\begin{figure}[thbp]
\centering
\includegraphics[width=1\linewidth]{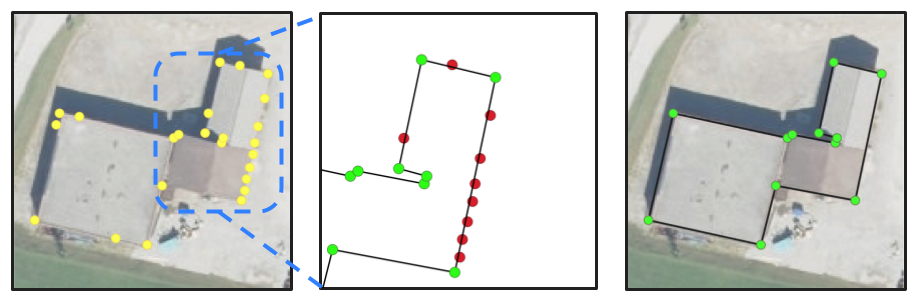}
\put (-246,72) {\scriptsize{prediction from CNN}}
\put (-161,72) {\scriptsize{ordering and filtering}}
\put (-77,72) {\scriptsize{final polygon}}
\caption{Polygon extraction steps: given the regularized building footprint, a CNN model detects all the building corners candidates (yellow vertices). The vertices are then sorted to produce a valid set of polygon coordinates. Points which lie too close to a building edge are filtered (in red). The final set of coordinates which describes the polygon is highlighted in green.}
\label{fig:mask2poly}
\end{figure}

Once the building footprints have been regularized, we extract a polygon for each building instance.

This task is accomplished using a simple \gls{gl:CNN} for corner detection.
The model receives the regularized mask as input and produces a corner proposal probability map.
Pixels with a value higher than a certain threshold in the probability map can be considered valid corners for the building polygon.

During inference each regularized footprint is evaluated by the corner detection network independently.
The detected points are then ordered clockwise moving along the perimeter of the regularized footprint in order to produce a valid set of coordinates for the polygon. 
As a final step, we filter redundant points that lie close to an edge as shown in ~\cref{fig:mask2poly}.

\section{Experiments}
\label{sec:Experiments}
\subsection{Experimental setup}
\subsubsection{Dataset}
The proposed pipeline has been evaluated on several aerial and satellite building segmentation datasets: INRIA~\cite{maggiori2017dataset}, CrowdAI~\cite{Mohanty:2018}, and SpaceNet~\cite{Urban3D2017}.

The INRIA dataset is an aerial dataset which covers a wide range of urban settlement appearances from different geographic locations. 
The particularity of this dataset is that the cities included in the test set are different from those of the training set, and it is composed of 180 training and 180 testing $5000 \times 5000$ orthorectified images with a resolution of 30 cm.
The CrowdAI dataset consists of 280,000 satellite images for training and 60,000 images for testing with an image resolution of $300 \times 300$ pixels.
During the test set inference over 500,000 building instances are extracted and regularized.
The SpaceNet dataset is composed of 30-50 cm pan-sharpened RGB satellite images from two cities in Florida: Jacksonville and Tampa.
The dataset is split into 62 images for the test set and 174 images for the training set.
The provided images have $2048 \times 2048$ pixels size.

All these datasets have a wide variety of buildings with different sizes, shapes and complexities that make the extraction of regularized polygons challenging.

\subsubsection{Network Architecture}
The \textbf{regularization network} has a residual autoencorder structure as shown in~\cref{fig:workflow}.
The encoders $E_G$ and $E_R$ are a sequence of $3 \times 3$ convolutional layers followed by batch normalization~\cite{ioffe2015batch} and $2 \times 2$ max-pooling layers.
After every down-sampling operation the number of convolutional filters is doubled, while the tensor size is halved.
The decoder $F$ is composed by a chain of 8 residual layers~\cite{he2016deep} followed by $3 \times 3$ convolutions, batch normalization layers and $2 \times 2$ up-sampling operations. 
Compared to the architecture proposed in \cite{paper1}, our encoders only have two pooling layers in order to keep trace of fine details of the input mask.
As shown in \cref{sec:Experiments}, this choice allows the decoder $F$ to reconstruct with more accuracy the buildings received as input and at the same time it can regularize them effectively, regardless their shape and complexity.
The discriminator $D$ shares the same layer combination of the encoders $E_G$ and $E_R$ but it has a deeper architecture, with 4 max-pooling operations in total.

For the \textbf{corner detection network} we just simply exploit the architectural model of the network $G$ used for the building regularization but using only 4 residual layers.

\begin{table*}[!htbp]
\centering
\begin{tabular}{lllllllllllll}
\cline{2-13}
 & \multicolumn{12}{c}{INRIA} \\ \cline{2-13} 
 & \multicolumn{2}{c|}{Bellingham} & \multicolumn{2}{c|}{Bloomington} & \multicolumn{2}{c|}{Innsbruck} & \multicolumn{2}{c|}{San Francisco} & \multicolumn{2}{c|}{Tyrol} & \multicolumn{2}{c}{Overall} \\ \cline{2-13} 
 & \multicolumn{1}{c|}{IoU} & \multicolumn{1}{c|}{Acc} & \multicolumn{1}{c|}{IoU} & \multicolumn{1}{c|}{Acc} & \multicolumn{1}{c|}{IoU} & \multicolumn{1}{c|}{Acc} & \multicolumn{1}{c|}{IoU} & \multicolumn{1}{c|}{Acc} & \multicolumn{1}{c|}{IoU} & \multicolumn{1}{c|}{Acc} & \multicolumn{1}{c|}{IoU} & \multicolumn{1}{c}{Acc} \\ \hline
\multicolumn{1}{l|}{R2UNet} & \multicolumn{1}{l|}{70.30} & \multicolumn{1}{l|}{\textbf{97.04}} & \multicolumn{1}{l|}{72.94} & \multicolumn{1}{l|}{\textbf{97.40}} & \multicolumn{1}{l|}{\textbf{73.48}} & \multicolumn{1}{l|}{\textbf{96.85}} & \multicolumn{1}{l|}{\textbf{76.29}} & \multicolumn{1}{l|}{\textbf{91.85}} & \multicolumn{1}{l|}{75.92} & \multicolumn{1}{l|}{\textbf{97.84}} & \multicolumn{1}{l|}{\textbf{74.57}} & \textbf{96.20} \\ \hline
\multicolumn{1}{l|}{\Citet{paper1}} & \multicolumn{1}{l|}{63.90} & \multicolumn{1}{l|}{96.37} & \multicolumn{1}{l|}{63.65} & \multicolumn{1}{l|}{96.51} & \multicolumn{1}{l|}{60.20} & \multicolumn{1}{l|}{95.23} & \multicolumn{1}{l|}{55.97} & \multicolumn{1}{l|}{84.60} & \multicolumn{1}{l|}{65.56} & \multicolumn{1}{l|}{96.88} & \multicolumn{1}{l|}{59.81} & 93.92 \\ \hline
\multicolumn{1}{l|}{Ours} & \multicolumn{1}{l|}{\textbf{70.36}} & \multicolumn{1}{l|}{96.99} & \multicolumn{1}{l|}{\textbf{73.01}} & \multicolumn{1}{l|}{97.36} & \multicolumn{1}{l|}{73.34} & \multicolumn{1}{l|}{96.77} & \multicolumn{1}{l|}{75.88} & \multicolumn{1}{l|}{91.55} & \multicolumn{1}{l|}{\textbf{76.15}} & \multicolumn{1}{l|}{\textbf{97.84}} & \multicolumn{1}{l|}{74.40} & 96.10 \\ \hline
\end{tabular}
\vspace{0.2cm}
\caption{Quantitative evaluation of building extraction and regularization results on the INRIA dataset. Scores are obtained by submissions of the predictions to https://project.inria.fr/aerialimagelabeling/.}
\label{tab:INRIA}
\end{table*}

\begin{table*}[]
\centering
\begin{tabular}{lcccccccccccc}
\cline{2-13}
 & \multicolumn{12}{c}{SpaceNet} \\ \cline{2-13} 
 & \multicolumn{4}{c|}{Jacksonville} & \multicolumn{4}{c|}{Tampa} & \multicolumn{4}{c}{Overall} \\ \cline{2-13} 
 & \multicolumn{2}{c|}{IoU} & \multicolumn{2}{c|}{Acc} & \multicolumn{2}{c|}{IoU} & \multicolumn{2}{c|}{Acc} & \multicolumn{2}{c|}{IoU} & \multicolumn{2}{c}{Acc} \\ \cline{2-13} 
 & \multicolumn{1}{c|}{$\mu$} & \multicolumn{1}{c|}{$\sigma$} & \multicolumn{1}{c|}{$\mu$} & \multicolumn{1}{c|}{$\sigma$} & \multicolumn{1}{c|}{$\mu$} & \multicolumn{1}{c|}{$\sigma$} & \multicolumn{1}{c|}{$\mu$} & \multicolumn{1}{c|}{$\sigma$} & \multicolumn{1}{c|}{$\mu$} & \multicolumn{1}{c|}{$\sigma$} & \multicolumn{1}{c|}{$\mu$} & $\sigma$ \\ \hline
\multicolumn{1}{l|}{R2UNet} & \multicolumn{1}{c|}{\textbf{72.85}} & \multicolumn{1}{c|}{7.077} & \multicolumn{1}{c|}{\textbf{96.54}} & \multicolumn{1}{c|}{1.105} & \multicolumn{1}{c|}{\textbf{70.74}} & \multicolumn{1}{c|}{6.056} & \multicolumn{1}{c|}{\textbf{94.90}} & \multicolumn{1}{c|}{1.219} & \multicolumn{1}{c|}{\textbf{71.80}} & \multicolumn{1}{c|}{6.670} & \multicolumn{1}{c|}{\textbf{95.75}} & 1.406 \\ \hline
\multicolumn{1}{l|}{\Citet{paper1}} & \multicolumn{1}{c|}{59.17} & \multicolumn{1}{c|}{5.348} & \multicolumn{1}{c|}{94.73} & \multicolumn{1}{c|}{1.693} & \multicolumn{1}{c|}{57.99} & \multicolumn{1}{c|}{6.892} & \multicolumn{1}{c|}{92.58} & \multicolumn{1}{c|}{2.317} & \multicolumn{1}{c|}{58.58} & \multicolumn{1}{c|}{6.197} & \multicolumn{1}{c|}{93.65} & 2.296 \\ \hline
\multicolumn{1}{l|}{Ours} & \multicolumn{1}{c|}{70.90} & \multicolumn{1}{c|}{7.551} & \multicolumn{1}{c|}{96.29} & \multicolumn{1}{c|}{1.169} & \multicolumn{1}{c|}{69.04} & \multicolumn{1}{c|}{6.587} & \multicolumn{1}{c|}{\textbf{94.90}} & \multicolumn{1}{c|}{1.286} & \multicolumn{1}{c|}{69.97} & \multicolumn{1}{c|}{7.146} & \multicolumn{1}{c|}{95.50} & 1.463 \\ \hline
\end{tabular}
\vspace{0.2cm}
\caption{Quantitative evaluation of building extraction and regularization results on the SpaceNet dataset}
\label{tab:SpaceNet}
\end{table*}

\begin{table*}[!htbp]
\centering
\begin{tabular}{l|c|c|cccc}
\hline
\multicolumn{3}{c|}{\textbf{Dataset}} & \multicolumn{4}{c}{CrowdAI} \\ \hline
\multicolumn{3}{c|}{\textbf{Method}} & \multicolumn{2}{c|}{IoU} & \multicolumn{2}{c}{Acc} \\ \hline
\textbf{Baseline} & \textbf{Regularization} & \textbf{Polygonization} & \multicolumn{1}{c|}{$\mu$} & \multicolumn{1}{c|}{$\sigma$} & \multicolumn{1}{c|}{$\mu$} & $\sigma$ \\ \hline
R2U-Net & - & - & 80.44 & 16.10 & 95.86 & 5.20 \\ \cline{1-3}
R2U-Net & \textit{Zorzi et al.} & - & 76.95 & 15.34 & 94.75 & 5.47 \\ \cline{1-3}
R2U-Net & Ours & - & 79.87 & 15.93 & 95.57 & 5.28 \\ \cline{1-3}
R2U-Net & \textit{Zorzi et al.} & Ours & 76.67 & 13.37 & 94.62 & 5.14 \\ \cline{1-3}
R2U-Net & Ours & Ours & 80.03 & 14.24 & 95.55 & 5.09 \\ \hline
Mask R-CNN & - & - & 73.22 & 17.84 & 94.38 & 4.77 \\ \cline{1-3}
Mask R-CNN & \textit{Zorzi et al.} & - & 71.72 & 17.32 & 93.88 & 4.82 \\ \cline{1-3}
Mask R-CNN & Ours & - & 73.57 & 17.65 & 94.34 & 4.74 \\ \cline{1-3}
Mask R-CNN & \textit{Zorzi et al.} & Ours & 72.13 & 13.82 & 92.57 & 4.80 \\ \cline{1-3}
Mask R-CNN & Ours & Ours & 74.23 & 14.51 & 94.12 & 4.75 \\ \hline
\end{tabular}
\vspace{0.2cm}
\caption{Quantitative evaluation of building regularization and polygonization results on the CrowdAI dataset}
\label{tab:CrowdAI}
\end{table*}

\subsubsection{Training Details}
Unlike the training approach proposed in \cite{paper1} where building instances are scaled and forward-propagated through the regularization network one by one, we train our \gls{gl:GAN} using $256 \times 256$ patches cropped from the dataset samples.
This helps to learn a generator and discriminator aware of the shape differences between small, medium and big buildings.
As ideal masks we exploit the accurate and good looking building footprints present in the ground truth of the chosen datasets.
The model is trained with batch size of 4 for 140,000 iterations.
We set $\alpha=3$, $\beta=1$, $\gamma=3$ in \cref{eq:full_objective}. 
$\epsilon$ and $\delta$ are kept to $0$ for the first 40,000 batches, then they are linearly increased to $1$ and $175$, respectively, in the following 40,000 batches to keep the learning more stable.
The weight matrix $W$ and $\hat{W}$ for \textit{Potts loss} and \textit{normalized cut loss} in the \cref{eq:regularized} are computed using the same expression and hyper-parameters described in \cite{paper1}.

Since the datasets we use for evaluation provide the ground truth already rasterized, the \gls{gl:CNN} used to detect building corners is trained using the building polygons available in OpenStreetMap for the cities of Chicago and Jacksonville.

For the initial building segmentation we used \gls{gl:R2UNet} trained with $448 \times 448$ patches randomly cropped from the SpaceNet and INRIA image samples.
In CrowdAI we directly train the model using the $300 \times 300$ images provided in the dataset.
Also, we provide some results using Mask R-CNN~\cite{he2017mask} as baseline using the pre-trained weights available in ~\cite{Mohanty:2018}.

During the training of all the networks, we applied standard data augmentation to the images (random rotations and flipping) and we trained all the pipeline models using Adam~\cite{kingma2014adam} optimizer with learning rate set to $0.0001$.

\subsection{Results}
\label{sec:Results}

In the \textbf{INRIA} and the \textbf{SpaceNet} datasets we compare against the baseline method and \citet{paper1}. 
The baseline exploits \gls{gl:R2UNet} as backbone to perform the initial building segmentation.
The results are then processed by the regularization method described in \cite{paper1} and by our building extraction method to produce the final footprints.
The final scores, based on \gls{gl:IoU} and accuracy, are shown in~\cref{tab:INRIA} and~\cref{tab:SpaceNet}.
Our building refinement can achieve quantitative results comparable or, in some test areas, even higher than the pure baseline.
Our approach, in fact, gets the higher \gls{gl:IoU} values in the test areas of Bellingham, Bloomington and Tyrol from the INRIA dataset and demonstrates to achieve accuracies very close to the pure baseline solution in the SpaceNet dataset.
This is a sign that the pipeline, made by multiple modules connected in cascade, does not lead to a significant drop in performance.
It is worth noting that the method \citet{paper1} has a significant \gls{gl:IoU} drop in these two datasets.
This is caused by the network architecture which is not capable to generalize well for big and complex buildings as shown in the results in~\cref{fig:results}.

In \textbf{CrowdAI} we test both \gls{gl:R2UNet} and Mask R-CNN as baseline networks for the initial segmentation.
Again, the proposed regularization can achieve results close to the pure segmentation network.
The \gls{gl:IoU} and accuracy scores achieved by \citet{paper1} are explainable considering that the CrowdAI dataset is mainly composed of midsize and small size constructions, with a low number of corners.

\subsubsection{Qualitative results}

\begin{figure}%
    \centering
    {{\includegraphics[width=0.45\linewidth]{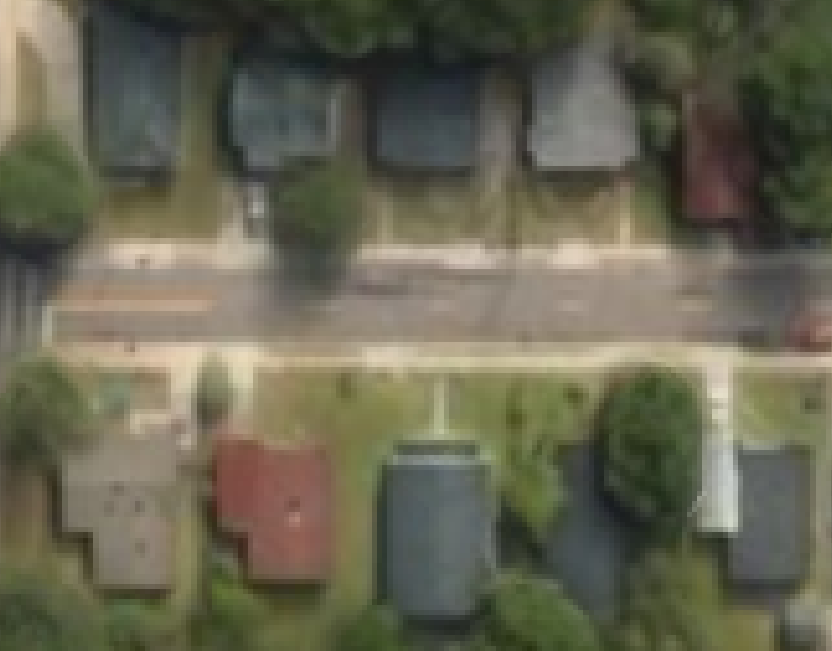} }}%
    {{\includegraphics[width=0.45\linewidth]{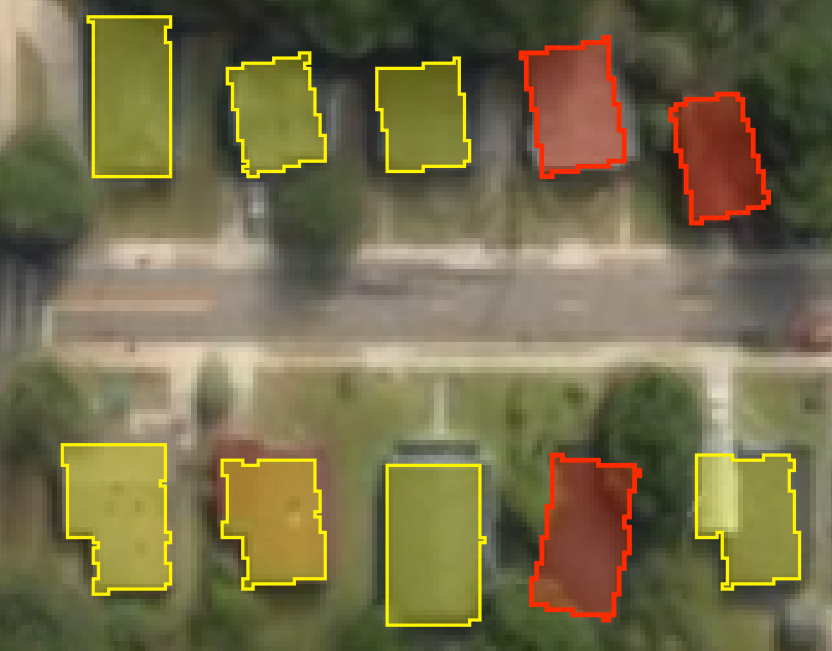} }}%
    \caption{On the left side: satellite image with occluded constructions. On the right side: result of the regularization network. Extracted footprints with wrong pose are highlighted in red.}%
    \label{fig:wrong}%
\end{figure}

\begin{figure*}
\centering
\begin{subfigure}[t]{\dimexpr0.19\textwidth+20pt\relax}
    \makebox[20pt]{\raisebox{40pt}{\rotatebox[origin=c]{90}{Inria bellingham}}}%
    \includegraphics[width=\dimexpr\linewidth-20pt\relax]
    {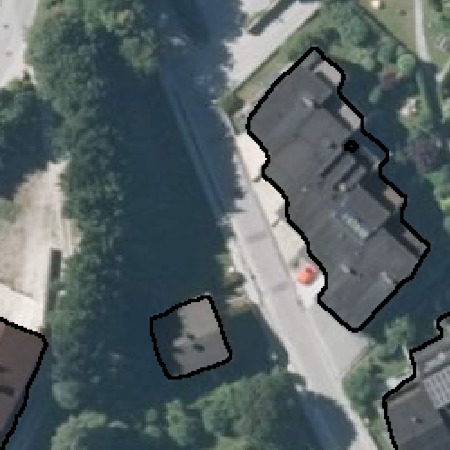}
    \makebox[20pt]{\raisebox{40pt}{\rotatebox[origin=c]{90}{Inria bloomington}}}%
    \includegraphics[width=\dimexpr\linewidth-20pt\relax]
    {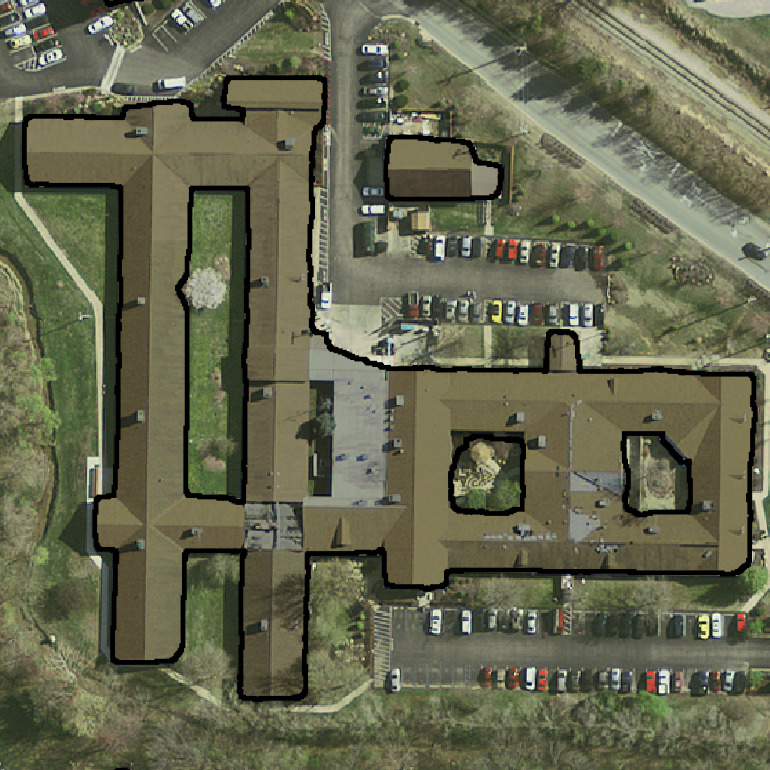}
    \makebox[20pt]{\raisebox{40pt}{\rotatebox[origin=c]{90}{CrowdAI}}}%
    \includegraphics[width=\dimexpr\linewidth-20pt\relax]
    {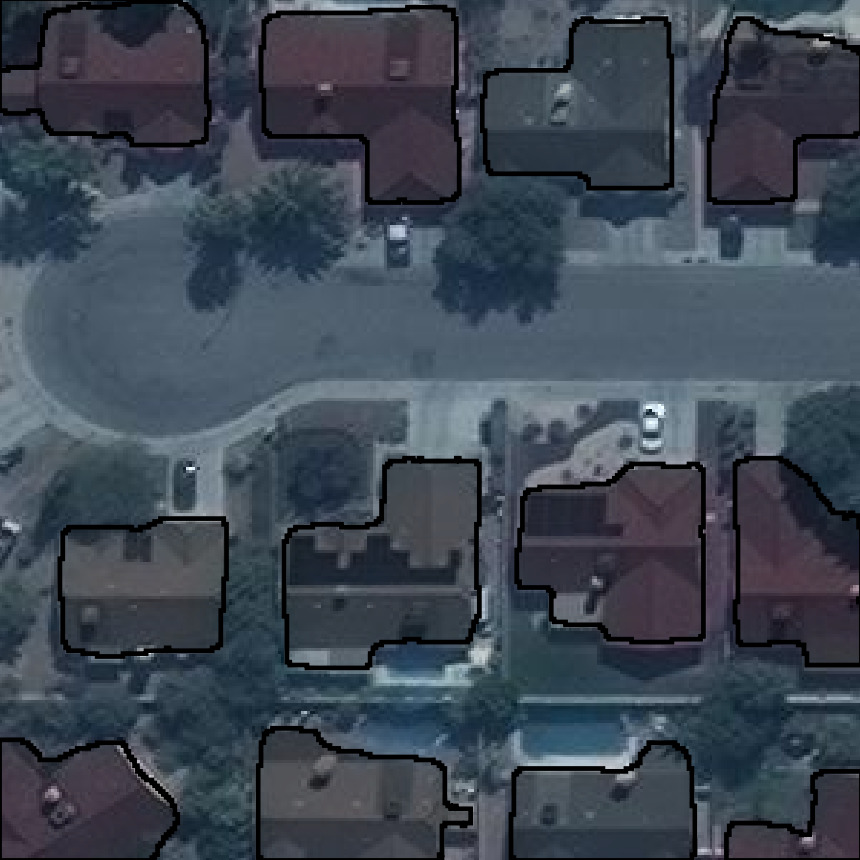}
    \makebox[20pt]{\raisebox{40pt}{\rotatebox[origin=c]{90}{SpaceNet}}}%
    \includegraphics[width=\dimexpr\linewidth-20pt\relax]
    {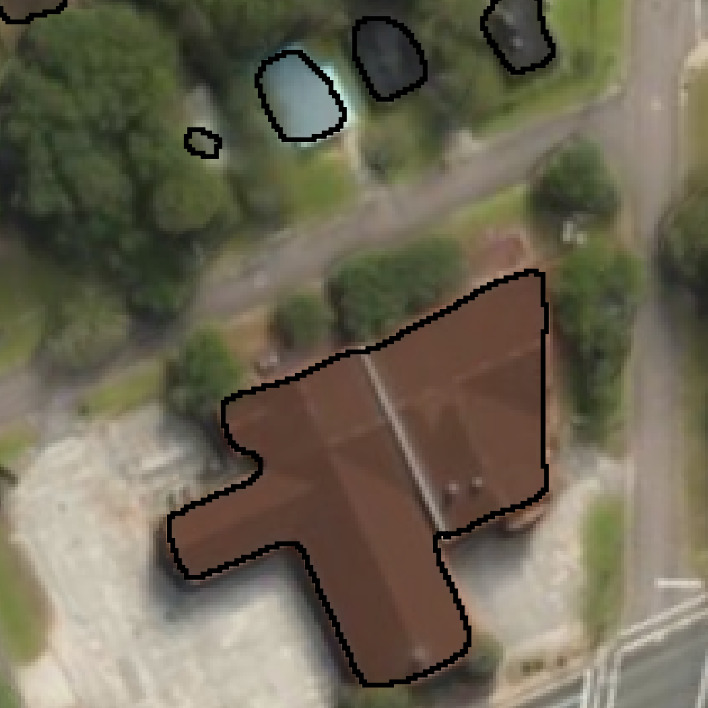}
    \caption{Segmentation}
\end{subfigure}
\begin{subfigure}[t]{0.19\textwidth}
    \includegraphics[width=\textwidth]  
    {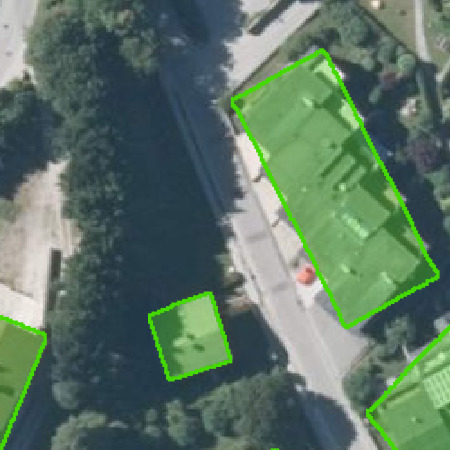}
    \includegraphics[width=\textwidth]
    {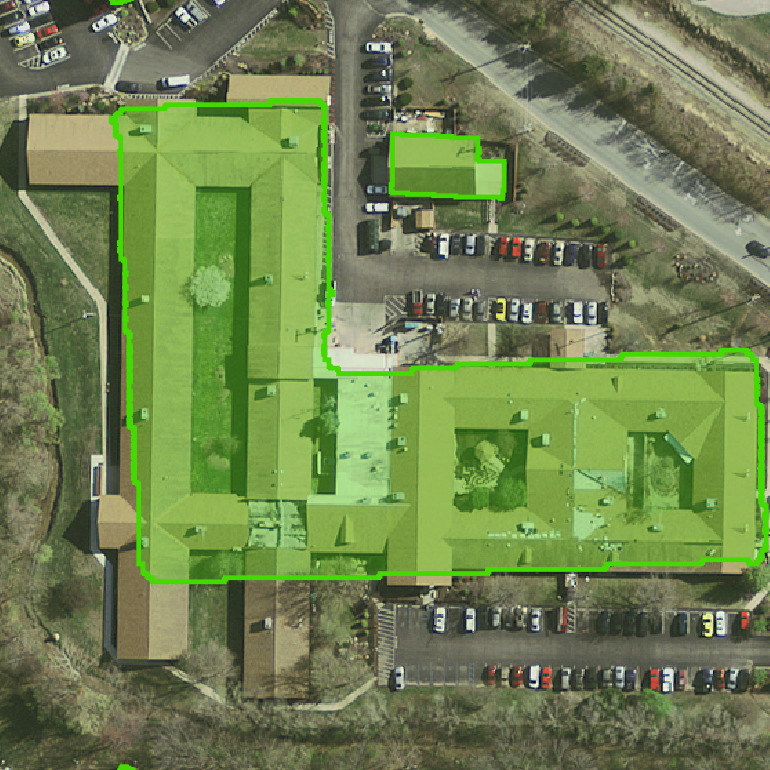}
    \includegraphics[width=\textwidth]
    {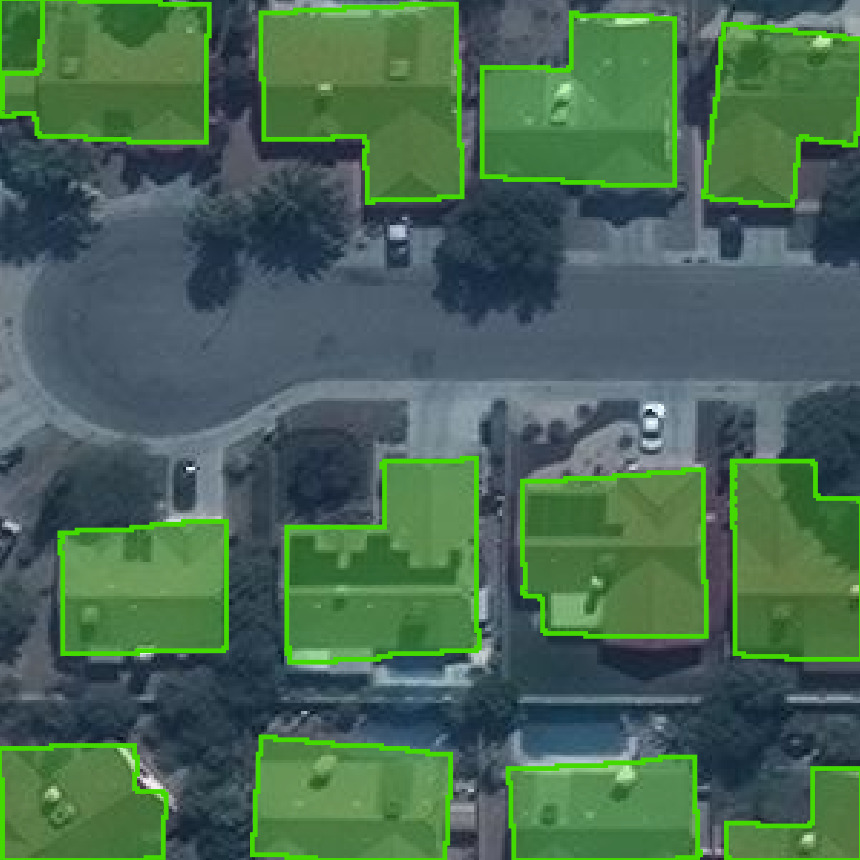}
    \includegraphics[width=\textwidth]
    {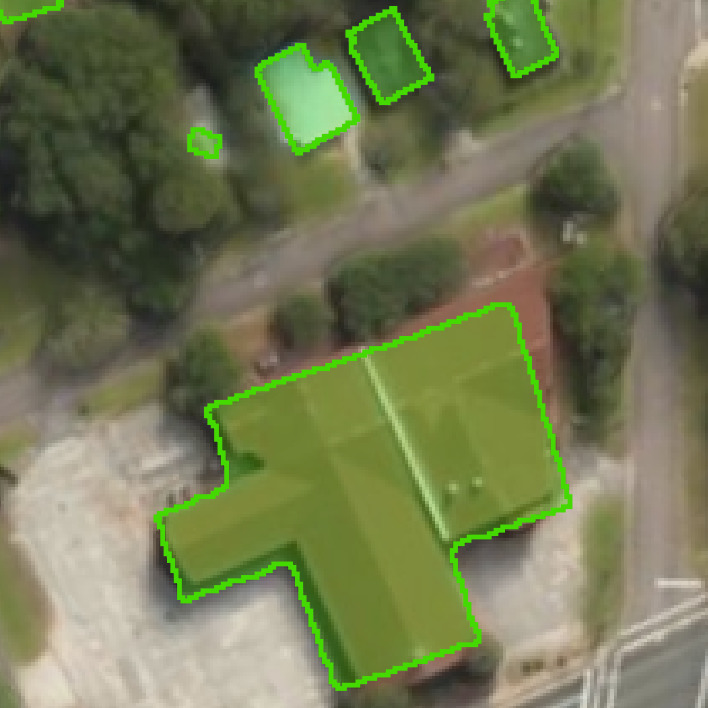}
    \caption{\Citet{paper1}}
\end{subfigure}
\begin{subfigure}[t]{0.19\textwidth}
    \includegraphics[width=\textwidth]  
    {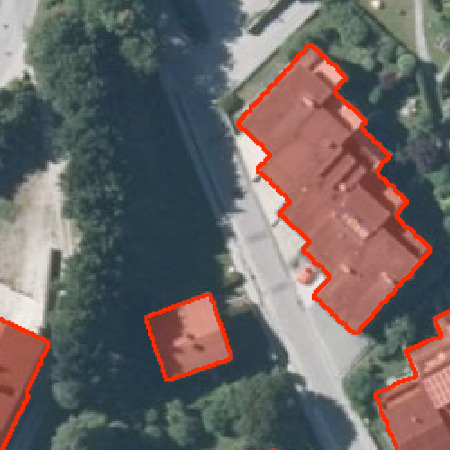}
    \includegraphics[width=\textwidth]
    {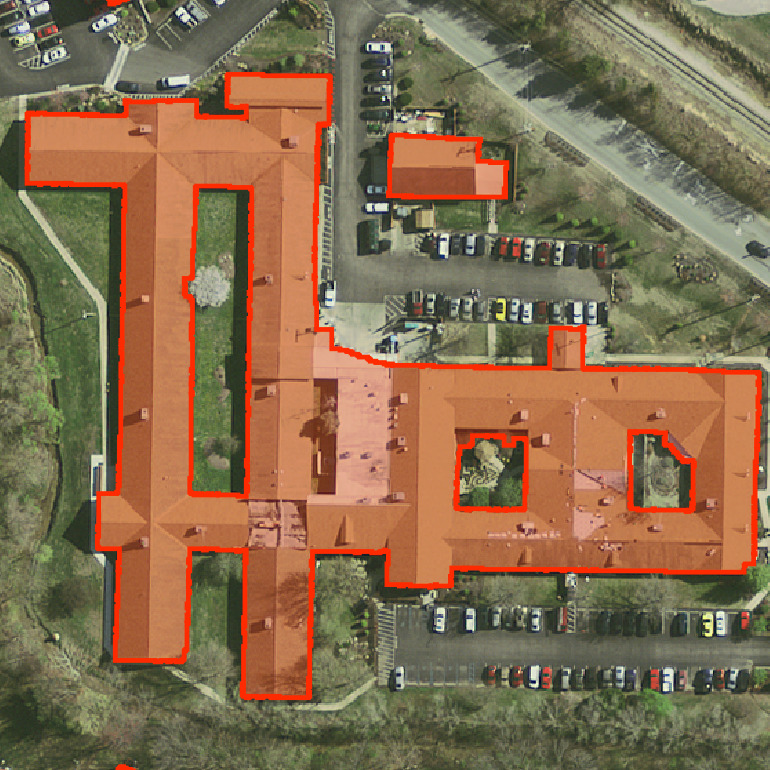}
    \includegraphics[width=\textwidth]
    {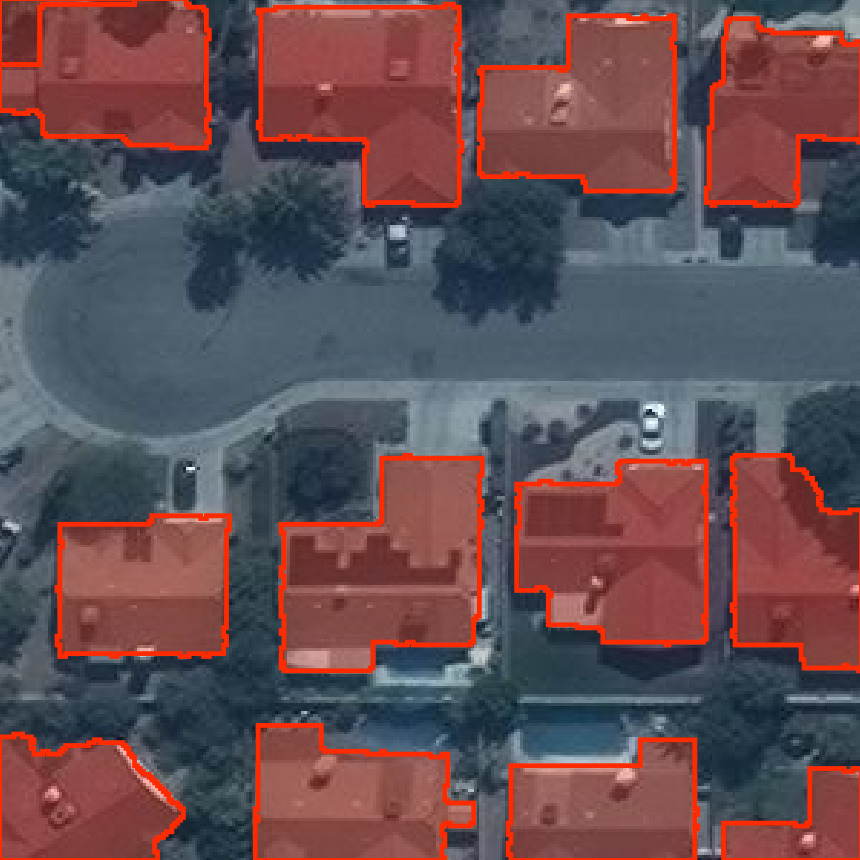}
    \includegraphics[width=\textwidth]
    {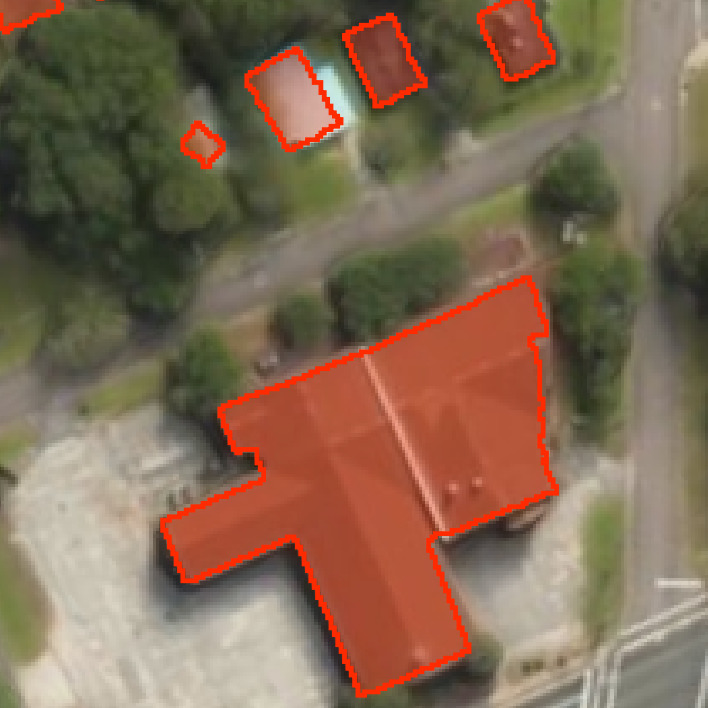}
    \caption{Proposed regularization}
\end{subfigure}
\begin{subfigure}[t]{0.19\textwidth}
    \includegraphics[width=\textwidth]  
    {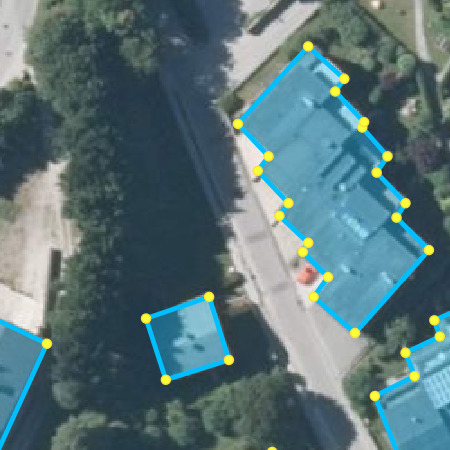}
    \includegraphics[width=\textwidth]
    {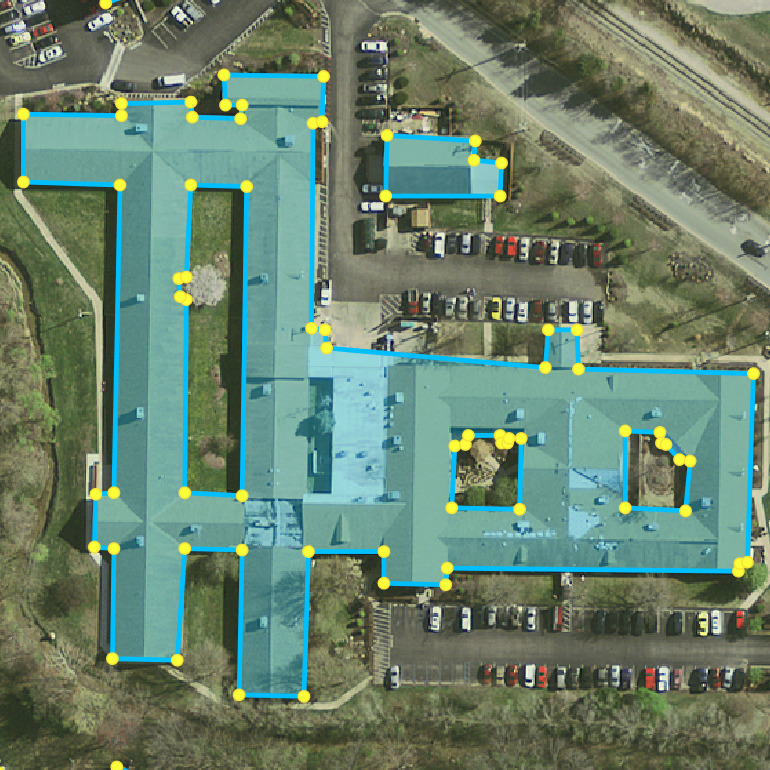}
    \includegraphics[width=\textwidth]
    {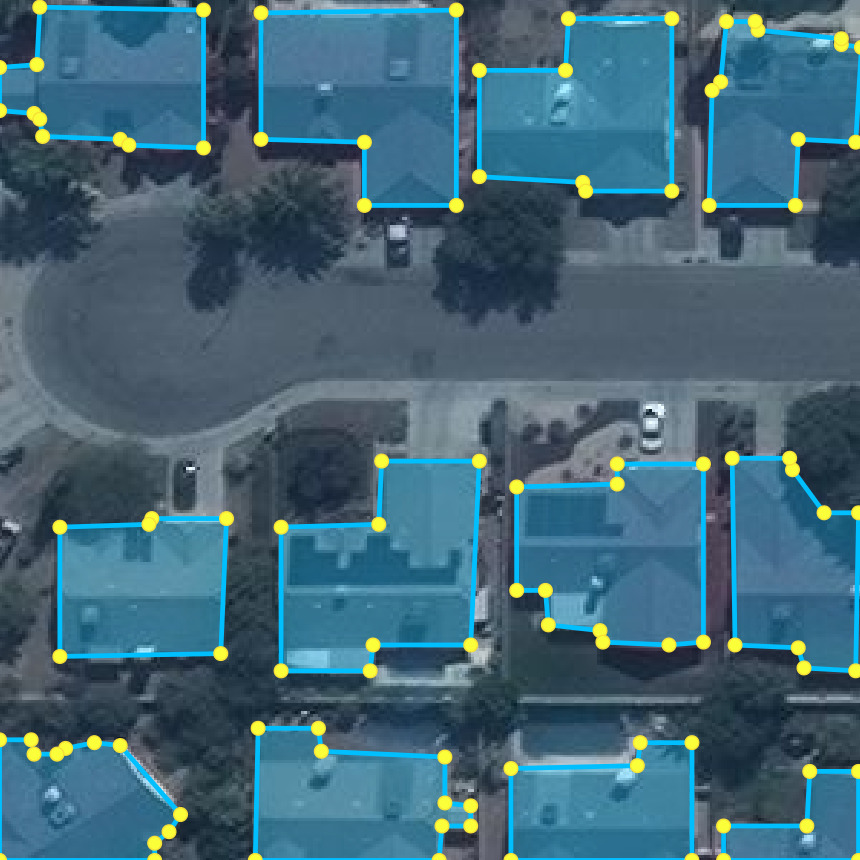}
    \includegraphics[width=\textwidth]
    {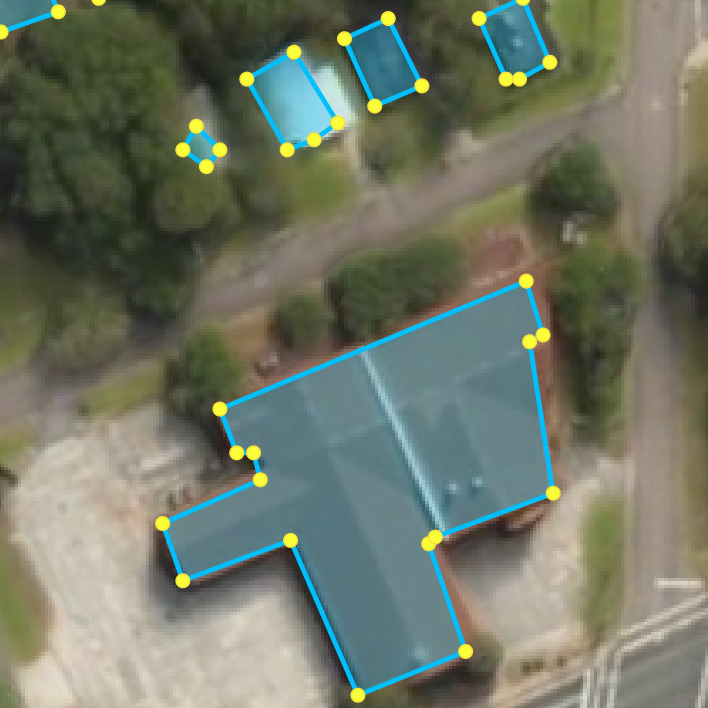}
    \caption{Extracted polygons}
\end{subfigure}
\caption{Buildings extraction results overlaid on top of a sample areas from Inria, CrowdAI and SpaceNet datasets.}
\label{fig:results}
\end{figure*}

We visualize some building footprints generated with different approaches in \cref{fig:results}.
Building footprints extracted with \cite{paper1} are accurate and visually pleasing if the building has a low number of vertices.
Vice versa, if the construction is complex, the network fails on producing a decent building boundary.

The algorithm proposed in this paper overcomes this problem producing accurate and realistic footprints regardless of the building size and complexity.
It is worth noting that our polygon extraction algorithm can also deal with inner courtyards creating a polygon for each building perimeter, as shown in the second row of \cref{fig:results}.

Despite the good results obtained in most of the circumstances, the proposed method is still not capable to extract sufficient context information to perform a correct regularization in the presence of occlusions.
In \cref{fig:wrong} is shown a residential area evaluated by \MaskPolydot.
The presence of the road in front of the constructions arranged in a line would suggest that the occluded buildings are also facing the street, in opposition with the extracted footprints.
Embedding a constraint about the disposition and the orientation of all the constructions in the scene would help the regularization network producing a coherent cartographic map of the the buildings from satellite or aerial images.

\glsresetall
\section{Conclusion}
\label{sec:Conclusion}

In this paper, we presented an approach for building segmentation and regularized polygon extraction, composed of three different and independent neural network modules.

The combination of the adversarial and the regularized losses results in a effective geometry constrain for the constructions, and encourages our predicted footprints to match building boundaries.
Furthermore, the regularization allows us to extract precise building polygons using a simple but effective \gls{gl:FCN} for corners detection.

The proposed method has proved to be capable not only of achieving equivalent or even higher results in terms of IoU and accuracy compared to state-of-the-art segmentation networks, but also of generating realistic and visually pleasing construction outlines that can be used in many cartographic and engineering applications.

\printbibliography

\end{document}